# The Achilles' Heel of AI: Fundamentals of Risk-Aware Training Data for High-Consequence Models

Dave Cook*[a], Tim Klawa[b]

[a]The Training Data Project, 4574 Indian Rock Terrace NW, Washington DC 20007-2567, [b]Figure Eight Federal, 1700 N Moore St #1840, Arlington, VA 22209

## ABSTRACT

AI systems deployed in high-consequence environments—such as defense, intelligence, and disaster response—must detect rare, high-impact events under operational constraints. Traditional annotation strategies that emphasize volume over value fail in these settings, introducing redundancy, noise, and risk. This paper presents **smart-sizing**, a strategic approach to training data curation that prioritizes informational value, label diversity, and model-guided selection. We introduce **Adaptive Label Optimization (ALO)** as an implementation framework for smart-sizing, combining pre-labeling, human-in-the-loop feedback, disagreement analysis, and marginal utility-based stopping rules.

Through empirical experiments, we show that models trained on only 20–40% of a curated dataset match or exceed the performance of full-data baselines, particularly in rare class recall and edge-case generalization. We also demonstrate how systematic labeling errors embedded in both training and validation sets can produce misleading model evaluations, underscoring the need for internal audit mechanisms and data governance.

Smart-sizing reframes annotation from a static task to a feedback-driven protocol aligned with operational goals. It offers a measurable path to improving AI robustness while reducing labeling cost, enabling teams to label what matters—and stop when it no longer helps.

## 1. INTRODUCTION

AI systems are falling short in the environments that matter most. In emerging defense use cases, research models struggle to detect obscured infrastructure, mobile threats, or improvised materials in contested terrain. Early ignitions go undetected in wildfire response when obscured by smoke or terrain anomalies. In humanitarian crises, makeshift shelters and disrupted infrastructure are often misclassified or ignored. These errors are not just technical. They are systemic. They reflect a fundamental gap in how training data is curated, prioritized, and aligned to mission needs.

The dominant mindset in AI development still assumes that more data will solve more problems. But in complex, high-consequence domains, that assumption breaks down. Larger datasets often introduce redundancy rather than insight. Labeling more of the same does not help models adapt to what is rare, ambiguous, or operationally significant.

This paper introduces smart-sizing, a methodology for constructing training datasets that are selective, performance-aligned, and consequence-aware. Smart-sizing is not about reduction for its own sake. It is about optimizing the selection, timing, and scope of labeled data to support model generalization where it matters most. It poses a critical question for modern AI programs: *How much data is enough, and which data is worth labeling in the first place?*

Training data should be treated as a strategic asset, not simply a prerequisite for model building. Labeling decisions must be integrated with risk management, cost control, and mission objectives—not driven by volume, uniformity, or completion metrics. In many AI programs today, annotation consumes the majority of development time and budget, often without feedback loops to assess impact. Over labeling inflates cost, embeds noise, and can reinforce a false sense of model confidence. Labeling for the common while ignoring the exceptional leads to brittle models that fail precisely at the margins—where missions succeed or fail.

*dave@trainingdataproject.org; phone 1 703-989-8409;

We propose Adaptive Label Optimization (ALO) as the operational framework for implementing smart-sizing. ALO integrates pre-labeling models, SME review, and dynamic metrics—such as disagreement and label diversity—to determine which data should be labeled next, and when labeling should stop. It treats model error not as failure, but as a diagnostic signal of what the model has yet to learn. In place of static quotas, ALO enables performance-informed **iteration** grounded in feedback and value.

Smart-sizing requires a fundamental shift from treating labels as interchangeable units to treating them as decision points that affect cost, trust, and operational readiness. This shift is especially critical in resource-constrained environments where compute is limited, labeling capacity is scarce, and system performance must be validated under real-world conditions. In these contexts, progress depends not on coverage but on insight**.**

This paper presents the architecture, implementation, and implications of ALO. We show how smart-sizing improves generalization, reduces labeling cost, and strengthens mission alignment. More fundamentally, we argue for a new way of thinking about training data: not as a backlog to be processed, but as intelligence to be engineered, reused, and governed**.**

## 2. TRAINING DATA AS A STRATEGIC VARIABLE: LESSONS FROM THE FIELD AND THE LITERATURE

The limitations of scale-centric annotation strategies are increasingly evident. While large datasets have driven many of the breakthroughs in deep learning, brute-force labeling without attention to quality, redundancy, or operational relevance has become a liability in high-consequence applications. In domains where models must perform reliably under uncertainty and in real-world complexity, the cost of mislabeling or over labeling is beyond inefficiency. It introduces risk that can cascade into operational failure.

The data-centric AI movement has emerged in response to this reality. Andrew Ng has called for a shift away from iterative model tuning and toward disciplined data curation. He argues that "50 thoughtfully engineered examples can be sufficient to explain to the neural network what you want it to learn," particularly in domains where data collection is expensive or high-reliability outcomes are essential. This framing challenges the common assumption that model performance improves in proportion to label volume.

Maggio et al. (2023) provided empirical evidence in support of this view. Across six public benchmarks, their experiments showed that models trained on just 30 percent of the data achieved over 95 percent of full-dataset accuracy. In some cases, using only 5 to 16 percent of the data yielded near-equivalent results, with training time reduced by over 100 times. However, their selection methods were largely based on random or uniform subsampling, not on mission-informed prioritization or performance-based selection.

Label quality compounds this challenge. Sambasivan et al. (2021) introduced the concept of data cascades, highlighting how early errors in annotation or schema can propagate through development, leading to brittle systems that pass validation but fail in deployment. Their research emphasized that annotation is not a one-time cost but a source of latent technical debt that grows without governance or feedback.

Kotian et al. (2023) proposed a partial remedy through Target-Aware Active Learning (TAAL), which prioritizes data based on class-specific performance gaps rather than model uncertainty alone. In TAAL, underperforming or low-recall classes receive targeted sampling until they reach predefined thresholds. This method moves annotation from being driven by data frequency to being guided by model performance.

More recent work has explored alternative strategies that further reduce reliance on exhaustive manual labeling. Nagase et al. (2025) developed a method for training object detectors without any new manual annotations by extracting pseudo-labels from multiple vision-language models such as GLIP and Grounding DINO. Their approach uses test-time augmentation and score calibration to refine pseudo-labels and integrate them into lightweight detectors. This allows detection systems to be deployed in new domains without fresh labeling efforts, although it requires confidence metrics and validation procedures to assess the integrity of the resulting labels.

Li et al. (2025) introduced F2SOD, a federated few-shot object detection framework that addresses data scarcity and privacy in edge environments. Their method enables distributed training of object detectors on small, locally held datasets by augmenting the data using diffusion models. The system allows clients to collaboratively refine models for novel classes while preserving data locality and security. Their results reinforce the argument that effective object detection does not require large-scale, centralized annotation if data diversity and transfer strategies are used intelligently.

In a complementary direction, Haider and Michahelles (2021) developed a semi-automatic labeling assistant that incorporates few-shot and one-shot detection models into annotation tools. Their system pre-labels images using learned representations, allowing human annotators to make quick corrections rather than labeling from scratch. They report labeling speeds two to six times faster than manual annotation, with equivalent quality. This approach exemplifies how machine assistance can reduce human workload without compromising integrity, but it still depends on having mechanisms in place to detect disagreement and prioritize edge cases.

Despite these innovations, the tools and workflows that manage large-scale labeling remain underdeveloped. Sager et al. (2021) reviewed annotation systems across computer vision and found that most focus on task execution speed or user experience rather than quality control or performance-linked feedback. Few tools enable annotators to understand where the model is uncertain, where disagreement is occurring, or where new labels would meaningfully improve generalization. These limitations are particularly critical in defense and humanitarian applications, where the cost of false positives or negatives is not just statistical but operational.

Label integrity is not just a training concern. Northcutt et al. (2021) found that benchmark test sets contain labeling errors at rates exceeding 3 percent on average, with some datasets performing substantially worse. Their analysis showed that high-capacity models trained on noisy data often appeared more accurate than smaller models because they learned to replicate systematic labeling flaws. This finding challenges the idea that validation performance is a reliable indicator of true model quality and underscores the need for internal audit mechanisms that track label error risk throughout development.

Our work builds directly on this research. The contributions of Ng, Maggio, Sambasivan, Kotian, Nagase, Li, Haider, Northcutt, and others have reshaped how the field thinks about labels—not as artifacts but as levers for performance, risk, and cost. What remains missing is a governed, system-level framework that translates these insights into operational practice.

Drawing from our experience supporting large-scale labeling efforts in the Department of Defense and Intelligence Community, we offer such a framework. Our approach integrates label value estimation, disagreement tracking, performance monitoring, and cost control into a unified system of annotation governance. Grounded in the smart-sizing doctrine and implemented through Adaptive Label Optimization, it provides teams with the ability to label what matters, stop when labeling no longer helps, and align training data with operational success.

### 3. WHY WE LABEL THE WRONG THINGS: TYPICAL BARRIERS TO SMART-SIZING

A lack of effort or intent rarely causes challenges with AI training data. Most annotation pipelines fail not because labeling teams are careless but because the systems governing their work are optimized for speed, consistency, and coverage rather than impact. These structural incentives are deeply embedded in how data operations are scoped, measured, and managed across commercial and mission settings.

The first structural barrier is the **throughput-centric design of annotation pipelines**. In many programs, success is defined by label counts, task completion metrics, or cost per labeled instance. These metrics reward uniform progress and penalize ambiguity. As a result, the easiest data is often labeled first, and annotation continues long after the model has saturated on core classes. There is rarely a mechanism to stop labeling when performance gains flatten, or to shift focus to underrepresented or high-risk categories. Annotation becomes a quota-driven process rather than a performance-informed one.

A second barrier is the **lack of performance-linked feedback in data operations**. Most labeling workflows operate independently from model evaluation cycles. Annotators receive little or no information about where the model is failing, what edge cases are confusing, or which inputs could improve generalization. Without that feedback, labelers prioritize clarity and speed over novelty and diversity. This disconnect makes it nearly impossible to surface the examples that matter most, which would expand the model's capacity to reason in the field.

A third constraint is **label schema rigidity**. Annotation taxonomies are typically fixed at the beginning of a project and are difficult to change once labeling begins. Even when operational requirements evolve, label definitions often remain static. This rigidity prevents teams from introducing new classes, revising definitions, or refining the level of granularity as new failure modes are discovered. As a result, important distinctions are flattened, and critical signals are lost. When ambiguity does arise, annotators are more likely to skip an example than to escalate it for schema revision.

Another structural issue is the **absence of a mission context in annotation design**. Many datasets used in high-consequence domains, such as disaster response or national security, are annotated without input from subject matter experts or field operators. This detachment leads to schemas and priorities that reflect general scene understanding rather than operational objectives. For example, labeling every vehicle on a road may improve object detection metrics, but fail to capture tactical relevance, such as if the car is abandoned, mobile, or structurally modified. Labeling at scale without context produces models that perform well on paper but fail under real-world stress.

Finally, there is often **no governance mechanism for assessing label value or redundancy**. Without tools to evaluate when additional labels stop improving the model, teams default to continuing annotation until the budget runs out or the queue is empty. Redundancy accumulates quietly, consuming resources and inflating datasets with examples that reinforce known patterns but do not challenge the model. The cost of this redundancy is not just financial. It hardens model assumptions and blinds systems to variation, particularly in edge conditions where resilience is needed most. These structural issues are not technical limitations. They are management decisions embedded in data infrastructure and workflow design. Addressing them requires more than introducing better tooling. It requires reframing how annotation is scoped, measured, and controlled.

Smart-sizing is a direct response to these barriers. It introduces dynamic prioritization, performance-informed stopping conditions, and schema flexibility into the annotation process. It reframes labeling from a fixed, linear task into a dynamic, iterative process governed by model feedback, mission relevance, and operational risk. The case for smart-sizing is not about doing less labeling. It is about doing the right labeling, at the right time, for the right reasons.

## 4. THE CASE FOR SMARTSIZING

Smart-sizing is a strategic framework for optimizing training data selection in environments where performance must be achieved under constraint. These constraints may include compute limits, budget ceilings, annotation fatigue, or simply the cost of failing to detect what matters. In mission-driven domains such as national security, emergency response, or contested geospatial operations, labeling everything is neither feasible nor defensible. Smart-sizing addresses this reality by providing a structured approach to identifying which labels matter, which do not, and when to stop.

Where traditional annotation pipelines are built for coverage, smart-sizing is built for consequence. It is not synonymous with compression, pruning, or random subsampling. It does not assume that value can be found after the fact by down selecting a large dataset. Instead, it guides the labeling process from the outset by asking a fundamental question: **Will this label help the model learn something it does not yet understand, and is that knowledge critical to the task at hand?**

This approach is operationalized through several core principles:
- **Labeling prioritization is based on mission value and model need.**
  Candidate samples are triaged according to expected impact, which may reflect model uncertainty, rarity, feature novelty, class imbalance, or operational criticality.
- **Iterative, performance-aware annotation.**
  Labels are applied in sprints. Each cycle includes ML-assisted pre-labels, human correction, and evaluation against performance targets. Annotation halts when returns diminish.

- **Dynamic schema evolution.**
  Ontologies and class definitions are refined in response to blind spots, edge cases, and misclassified anomalies. This ensures the label space evolves with operational realities.
- **Integrated quality controls.**
  Labeling linters and model disagreement metrics guide annotators in real time, elevating ambiguous or high-disagreement cases for SME review. This creates feedback loops that preserve quality while enabling scale.
- **Stopping rules based on marginal value.**
  Annotation ceases not at dataset completion but when further labeling no longer yields measurable gains.

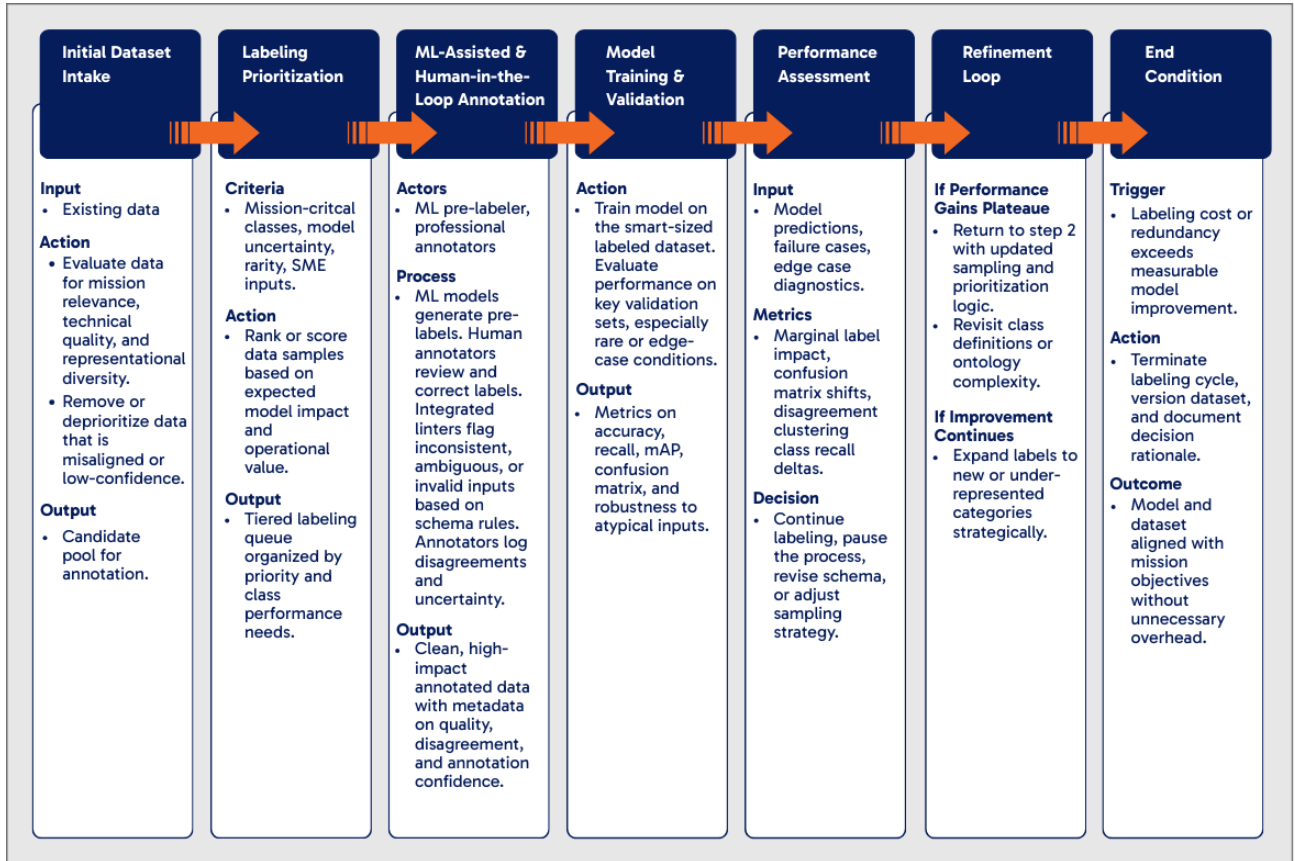

Figure 1: Smart-sizing Workflow for Strategic Labeling

To formalize this process, smart-sizing can be governed by a decision rule:

$$\text{Label if: } \frac{\Delta\text{Perf}(x)}{C(x)} > \tau$$

Where:
- $\Delta\text{Perf}(x)$ represents the **expected improvement** in model performance from labeling sample $x$, which may be derived from disagreement, class rarity, or model uncertainty.
- $C(x)$ reflects the **cost or complexity** of labeling $x$, such as SME time, annotation risk, or ambiguity.
- $\tau$ is a **minimum performance-to-cost threshold**, adjustable based on operational needs or labeling budget.

This inequality serves as a governing condition: **Label only when the informational return justifies the resource investment.** When $\Delta\text{Perf}(x)$ drops or $C(x)$ rises, the system adapts or stops.

Smart-sizing, then, is not about labeling less. It is about labeling with purpose. It converts annotation from a linear, coverage-focused activity into a strategic process governed by feedback, consequence, and control.

## 5. EXECUTING SMARTSIZING: THE ADAPTIVE LABEL OPTIMIZATION (ALO) METHOD

Smart-sizing provides the strategy; Adaptive Label Optimization (ALO) delivers the method. ALO is a structured, iterative process for conducting training data curation in environments where performance matters most and labeling resources are constrained. It enables model-directed annotation by integrating machine prediction, human judgment, and performance feedback into a continuous decision loop.

Each ALO cycle functions as a labeling sprint. The sprint is governed by evidence—how well the model is performing, what it is still failing to detect, and whether new labels are likely to improve that capability. ALO replaces static annotation with a process of intentional triage and control.

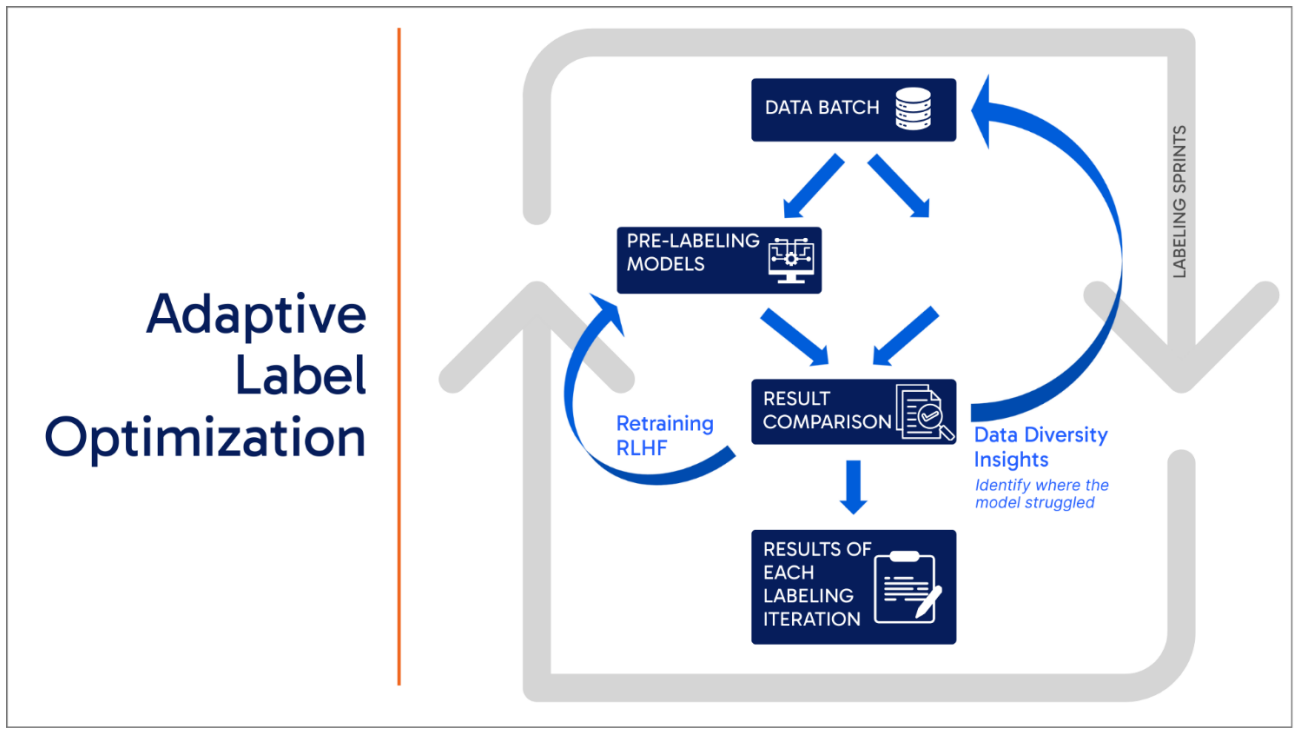

*Figure 2: The ALO Iterative Labeling Workflow*

### 5.1 ALO Workflow Stages

*1. Data Batch*

A curated set of candidate imagery is prepared for annotation. These images are filtered for operational relevance and quality, and may include recent collections, simulated edge cases, or archived imagery flagged during model failure reviews.

*2. Pre-labeling Models*

The current model is used to generate predicted labels. These pre-labels accelerate human annotation and act as indicators of confidence or confusion. Samples with uncertain predictions or misaligned class boundaries are prioritized for review.

*3. Subject Domain Experts*

Samples identified as ambiguous, novel, or high-risk are routed to SMEs. Experts refine labels, annotate complex cases, and may suggest revisions to class definitions. Their contributions ensure operational accuracy, especially in low-frequency or adversarial conditions.

*4. Result Comparison*

Human-reviewed labels and pre-label outputs are compared. Metrics such as disagreement rate, false positive/negative patterns, and cross-class confusion inform performance diagnostics and schema refinement.

*5. Retraining and Feedback Integration*

The model is retrained with updated annotations. Validation is conducted not just on aggregate metrics, but also on edge case performance, recall for rare classes, and robustness under visual degradation. Embedding space shifts are monitored to ensure diversity and learning progression.

*6. Data Diversity Insights*

Post-retraining analysis examines whether the new labels expanded model understanding. Metrics include embedding distribution spread, new cluster formation, and improved separation of high-confusion classes. These insights inform whether the next round of annotation is justified.

**5.2 Operationalizing the Labeling Decision: A Defense Use Case**

ALO's decision loop is governed by the smart-sizing principle introduced earlier:

$$\text{Label if: } \frac{\Delta\text{Perf}(x)}{C(x)} > \tau$$

To illustrate, consider an EO imagery task focused on curating a dataset for rare aircraft detection. The objective is to improve a model's ability to recognize unconventional or region-specific airframes under variable conditions.

- $\Delta\text{Perf}(x)$ is estimated using:
    - Model uncertainty (entropy of predictions)
    - Annotator-model disagreement
    - Edge case indicators (e.g., aircraft geometry, atypical configurations)
    - Prior underperformance on similar examples
- $C(x)$ includes:
    - Label complexity due to occlusion or poor resolution
    - SME time to resolve nonstandard configurations
    - Re-labeling cost from unclear schema fit

For example, an image containing a rare surveillance aircraft partially obscured on a desert airstrip, misclassified by the model as a generic cargo plane, would score high in $\Delta\text{Perf}(x)$. Given the operational relevance and model confusion, and despite the SME time required, the sample would be prioritized. Conversely, a well-lit image of a commercial jetliner in a class the model already handles would fall below the labeling threshold and be deferred or excluded.

This operational decision rule ensures that SME time and compute cycles are allocated not to what is easiest, but to what is most consequential.

**5.3 Closing the Loop**

Each ALO cycle concludes with an assessment of label utility. If rare class recall improves, if disagreement rates decline, and if feature diversity expands, the cycle proceeds. If progress stalls, labeling pauses, schema adjustments are made, or task focus is realigned.

ALO does not aim to label faster. It aims to label smarter. It turns annotation into a disciplined feedback system—one where each label earns its place by showing it matters.

# 6. WHAT THE DATA REALLY TELLS US: EVIDENCE FOR A SMARTER LABELING STRATEGY

The smart-sizing framework is built on the premise that training data should be evaluated and managed based on informational value, not volume. To validate this, we conducted a series of experiments focused on two key questions: How does label quality shape model performance and risk? And how much of a full dataset is truly needed to build a capable model? Our findings reinforce some conclusions from recent literature—but they also reveal new risks and challenge prevailing assumptions in both industry and academia. We provide empirical evidence that confirms the diminishing returns of redundant labels and exposes how systematic errors in annotation can distort validation, mask brittleness, and degrade downstream utility.

## 6.1 Label Quality and the Illusion of Model Performance

In the first set of experiments, we explored how systematic labeling errors affect model accuracy. We introduced artificial errors into a supervised training dataset at varying levels and tracked performance metrics over time. As expected, performance degraded with increasing error, but the effects were not linear or immediately obvious.

When the same labeling errors were applied to both training and validation sets, models appeared to perform well—even when up to 25 percent of the labels were incorrect. This created a false sense of model readiness. In reality, these models were learning and validating on the same flawed patterns. This experiment illustrates a key failure mode: systematic error propagation. Without safeguards during labeling, false confidence becomes embedded in model evaluation. Labeling flaws may go undetected until a model is deployed in the real world and fails under unfamiliar conditions.

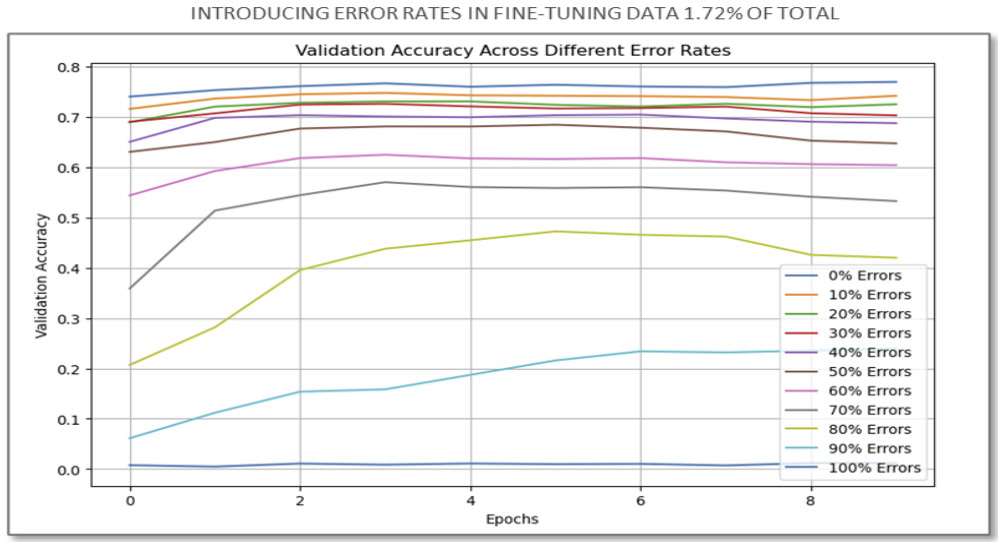

*Figure 3: Effect of Label Error on Model Accuracy*

Figure 3 shows validation accuracy trends for models trained on datasets with increasing levels of artificial labeling error. Despite 25% of labels being incorrect, the model appears to maintain strong validation performance, demonstrating how systematic label flaws can distort evaluation metrics and conceal true generalization loss.

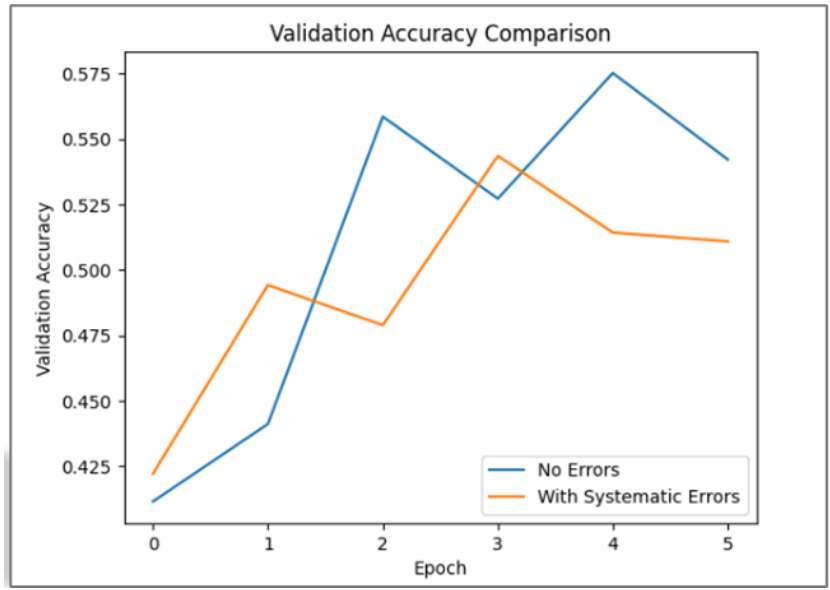

*Figure 4: Accuracy Distortion from Systematic Validation Errors*

Figure 4 shows a side-by-side comparison of models trained and validated on clean versus systematically flawed labels. Despite degraded true performance, validation curves remain elevated when both training and validation data share the same errors—highlighting how systematic label flaws can create misleading accuracy signals. This aligns with the observations of Sambasivan et al. (2021), but our work goes further by quantifying how validation performance can be systematically inflated by embedded error patterns—turning model evaluation itself into a misleading signal. These findings underscore a central risk in traditional AI workflows: label quality is often invisible until models are deployed, at which point remediation is difficult, expensive, or operationally infeasible.

**6.2 Label Quantity: Confirming and Challenging Assumptions About Scale**

In a separate line of experiments, we examined how much labeled data is truly necessary to train a performant model. The findings of **Maggio et al. (2023)** provide a starting point here: their work showed that models trained on about 30 percent of available data often achieved 95 percent of the performance of models trained on full datasets.

We found similar, but sharper, results. By curating subsets of data based on embedding space diversity and model disagreement (principles central to smart-sizing), we trained models on just 20 percent of the total dataset and achieved performance within 5 percent of full-dataset baselines. In some configurations, these smart-sized models even outperformed full-data models on rare class recall and edge case generalization.

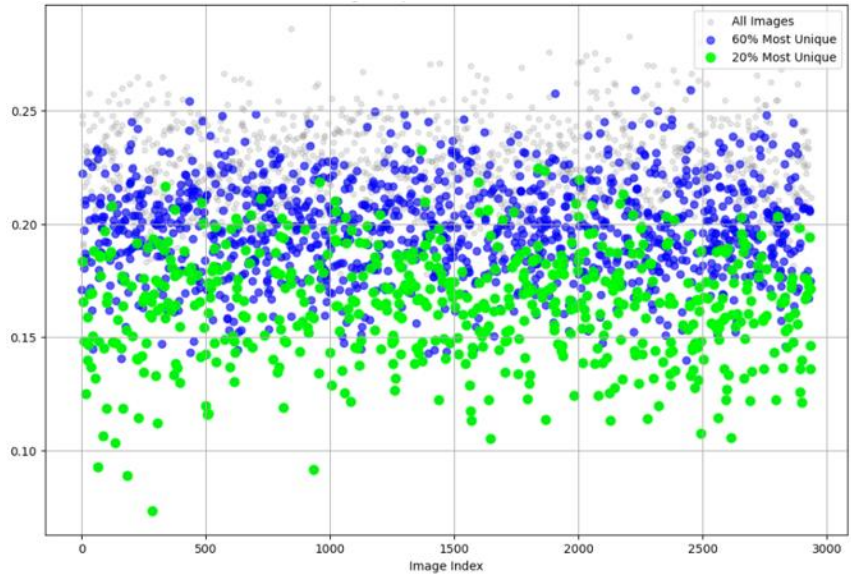

*Figure 5: Label Uniqueness Across Full and Subset Data*

Figure 5 shows samples ranked by embedding-based uniqueness scores. Green and blue points represent the top 20% and 60% most unique samples, respectively, while gray denotes the full dataset. The figure illustrates how smart-sizing prioritizes data with higher representational diversity to maximize model learning efficiency.

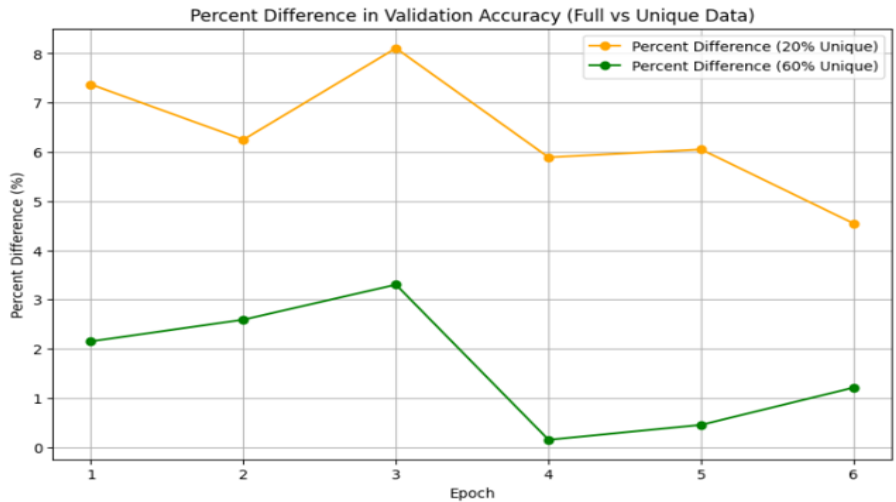

*Figure 6: Validation Accuracy Delta for Unique Subsets*

Figure 6 shows the percent difference in validation accuracy over six training epochs between models trained on the full dataset and those trained on the top 20% and 60% most unique samples. Despite a significantly smaller training set, the smart subsets maintain comparable performance, demonstrating the efficiency of high-value label selection.

This reinforces **Ng's argument for data-centric AI**: that quality and representational variety matter more than scale. However, our experiments move the discussion forward by showing how strategic selection—not just reduction—can preserve or enhance performance. Smart-sizing is not merely a leaner variant of traditional labeling. It is a fundamentally different logic that assigns value to each example and uses that value to inform action.

### 6.3 What Changes

Together, these findings shift the conversation in three ways:
1. **Validation metrics cannot be trusted without understanding label quality.** Without embedded QA in the annotation loop, models may learn and validate on flawed assumptions, reinforcing error patterns and inflating confidence. This risk is largely absent from mainstream workflows and tooling today.
2. **Redundancy is not neutral—it is a form of bias.** When frequent classes or standard scenes dominate the label set, models become insensitive to variation. Our results show that sampling for diversity improves generalization, reduces noise, and accelerates learning—especially in constrained settings.
3. **The "label more" doctrine is obsolete.** Smart subset selection, informed by disagreement, rarity, and mission relevance, can produce models that are not only more efficient to train but more aligned with operational success.

### 6.4 Operational Interpretation of Results

These findings also provide strong validation for the smart-sizing decision equation introduced earlier:

$$\text{Label if: } \frac{\Delta \text{Perf}(x)}{C(x)} > \tau$$

Our experiments show that the expected gain from labeling a sample, $\Delta\text{Perf}(x)$, is maximized when data is selected for diversity, novelty, or class rarity. Figures 3 and 4 illustrate how such samples contributed disproportionately to model improvement, even at small scale. Conversely, Figures 1 and 2 demonstrate that labeling cost, $C(x)$, includes not only human effort but also the downstream risk of embedding error into both training and validation sets. When label quality drops or redundancy increases, $C(x)$ rises and $\Delta\text{Perf}(x)$ flattens—violating the smartsizing threshold $\tau$ and signaling the need to pause, pivot, or stop labeling altogether.

This decision logic is no longer a theory. It is a quantifiable method, validated by experimental evidence and aligned with real-world operational needs.

### 7. FROM PIPELINE TO PROTOCOL: SMARTSIZING AS A STRATEGIC SHIFT IN AI DEVELOPMENT

The findings in this paper point to a larger transformation underway in machine learning development. Labeling is no longer a backend task. It is a performance-governing function that shapes every downstream decision. In light of emerging practices and technologies—ranging from annotation-free object detection to federated few-shot learning—the need for structured, risk-aware annotation governance has never been more urgent.

### 7.1 Reframing the Labeling Function

Our experiments and the broader literature demonstrate that model performance is governed less by the size of a dataset and more by the informational value and integrity of each labeled example. Recent advances confirm this across technical domains. Nagase et al. (2025) showed that it is possible to train object detectors without new human labels by leveraging pre-trained vision-language models. Their framework, while powerful, raises new questions about pseudo-label trust and the absence of human oversight in selecting meaningful examples.

Similarly, Li et al. (2025) demonstrated that few-shot object detection can be performed across federated clients using augmented novel data and collaborative fine-tuning. Their results confirmed that a small number of strategically chosen labels, supported by generative augmentation, can outperform larger, less targeted training sets. These findings validate smart-sizing's core claim: the value of a label is contextual and should be governed, not assumed.

Haider and Michahelles (2021) add to this picture by showing how pre-labeling and one-shot detection can accelerate human annotation with minimal loss in quality. Their results suggest a future in which annotation is neither manual nor automatic, but collaborative. However, their system, like others, lacks a formal structure for identifying when new labels stop contributing to performance, or how to prioritize ambiguous or rare cases under mission constraints.

### 7.2 Managing Cost and Complexity Through Intentional Design

Smart-sizing provides this structure. It enables teams to define performance thresholds, measure label utility in real time, and enforce stopping rules based on marginal value. This transforms annotation from a throughput activity into a managed, evidence-based protocol. In doing so, it allows for tighter control over cost, SME bandwidth, and compute cycles. It also limits the risk of overfitting redundant or flawed data—an increasingly common problem as model architectures scale but governance frameworks lag behind.

Smart-sizing also changes the tools and metrics that matter. Instead of label throughput or aggregate accuracy, teams using Adaptive Label Optimization can monitor:

- Marginal gain per label class
- Rare class recall
- Disagreement density
- Information uniqueness in embedding space

These are not merely technical metrics. They are signals of model resilience and mission readiness.

### 7.3 Organizational Implications

Finally, smart-sizing has implications for how teams are organized. Annotation becomes a cross-functional process involving model developers, SME reviewers, data managers, and program leads. This mirrors recent shifts toward distributed collaboration seen in federated learning and edge-based retraining. The challenge now is not just to reduce annotation—but to coordinate it intentionally, with oversight.

Smart-sizing enables this coordination. It treats each label not as a task to complete, but as a decision to justify. In operational terms, it is the equivalent of moving from pipeline to protocol—from execution to governance.

## 8. CONCLUSION

This paper has presented smart-sizing as a practical and strategic response to the failure modes of large-scale annotation in high-stakes AI. We began by reviewing a growing body of research that challenges the prevailing assumptions behind dataset size, label uniformity, and throughput-centric annotation. We then introduced **Adaptive Label Optimization (ALO)** as an implementation framework for smart-sizing—one that integrates model feedback, human-machine collaboration, and cost-governed stopping conditions into a repeatable and measurable labeling protocol.

Our experiments demonstrated that:
- Models trained on strategically selected subsets (as small as 20%) can match or exceed the performance of full-data baselines.
- Systematic labeling errors embedded in both training and validation datasets can distort model performance evaluation, producing artificial confidence that breaks down at deployment.
- Labeling quality, diversity, and disagreement are not merely data engineering issues; they are primary drivers of model risk.

We connected these findings to a broader transformation in the field: from data as fuel, to data as a controlled variable. Across new approaches in vision-language modeling, federated few-shot detection, semi-automatic annotation tooling, and benchmark audits, the field is converging on a realization: **we need not only smarter models, but smarter strategies for deciding what to teach them.**

Smart-sizing, as a doctrine, answers that need. ALO provides the mechanisms to implement it.

Looking ahead, several lines of inquiry remain open and essential:

**Formalizing Label Value Estimation:** While ALO relies on marginal gain metrics, the field lacks standardized methods for quantifying label informativeness across architectures and domains. Further research could develop generalized scoring functions, including probabilistic and uncertainty-aware formulations.

**Governance Tools and Interfaces:** Existing labeling tools do not support smart-sizing workflows natively. Future work should focus on building interfaces that allow real-time visualization of disagreement, embedding diversity, marginal label value, and model performance deltas—ideally integrated with annotation environments.

**Adaptation Across Modalities:** While this paper focused on computer vision, smart-sizing should be extended to language, audio, multi-modal models, and temporal data. Each domain presents unique challenges in data sparsity, ambiguity, and feedback loop timing.

**Human Factors and SME Utilization:** As ALO introduces targeted use of SMEs for ambiguous or rare-class samples, further work is needed to study how expert time is best allocated, and how tools can reduce SME fatigue, maintain quality, and align label schema with evolving mission requirements.

**Real-World Policy and Lifecycle Integration:** In defense, humanitarian, and infrastructure AI systems, training data evolves over months or years. Smart-sizing must be integrated with larger program lifecycle controls. Future research should explore how to couple ALO with data versioning, model retraining triggers, and institutional oversight.

This work represents a step toward a more rigorous discipline of **training data governance**. In the past, annotation was treated as a static precursor to modeling. Our results suggest it is better understood as an ongoing decision loop—where labeling becomes an act of strategy, not scale.

Smart-sizing is a call to act with greater precision. The future of AI will be shaped not only by what models can learn, but by how deliberately we choose to teach them. Our aim is to help build systems, tools, and policies that make such deliberate teaching possible.